\documentclass[10pt,twocolumn,letterpaper]{article}
\ifdefined\pdfminorversion\pdfminorversion=7\fi

\PassOptionsToPackage{table}{xcolor}
\usepackage[pagenumbers]{cvpr}
\usepackage{xcolor}
\usepackage{booktabs}
\usepackage{makecell}
\usepackage{multirow}
\usepackage{tabularx}
\usepackage{balance}
\usepackage{array}
\IfFileExists{preamble.tex}{

}{}

\definecolor{cvprblue}{rgb}{0.21,0.49,0.74}
\usepackage[pagebackref,breaklinks,colorlinks,allcolors=cvprblue]{hyperref}

\title{When Guidance Becomes a Guard:\\ Early Interception of Unsafe Inputs in Text-to-Image Generation}

\author{%
\begin{tabular}{c}
Yanchun Li\textsuperscript{1} \quad
Yujin Wu\textsuperscript{1} \quad
Gangkai Shen\textsuperscript{1} \quad
Ming Yang\textsuperscript{1} \\[0.4em]
Li Zeng\textsuperscript{2} \quad
Xianlong Wang\textsuperscript{3} \quad
Hangtao Zhang\textsuperscript{4} \quad
Shujuan Tian\textsuperscript{1} \\[1em]
{
\textsuperscript{1}Xiangtan University
\qquad
\textsuperscript{2}Changsha University of Science and Technology
} \\[0.3em]
{
\textsuperscript{3}City University of Hong Kong
\qquad
\textsuperscript{4}Huazhong University of Science and Technology
}
\end{tabular}%
}

\begin{document}
\maketitle

\begin{abstract}
Text-to-image diffusion models can produce unsafe content,
while output filtering intervenes only after synthesis.
We investigate whether native generation signals enable
early detection of unsafe and adversarial inputs.
We find that classifier-free guidance exposes strong safety
signals in conditional responses, accessible through
compact statistics and linear readouts.
Based on this finding, we propose Conditional Semantic
Trajectory Interception (CSTI), which aggregates activation
contrasts between conditional and unconditional responses
at early checkpoints for calibrated risk detection.
CSTI fits concept-specific detectors using canonical concept
and safe examples, without attack samples, intermediate
image decoding, or generator modification.
Across five generators, four adversarial attack collections,
and two general NSFW datasets, CSTI achieves the strongest
overall performance among six baselines at a matched 5\% FPR.
Early interception reduces generation latency by over
60\% on intercepted paths.
These results demonstrate the effectiveness of native
conditional responses for early safety interception.
\end{abstract}

\section{Introduction}
\label{sec:intro}

\begin{figure*}[t]
    \centering
    \includegraphics[width=\textwidth]{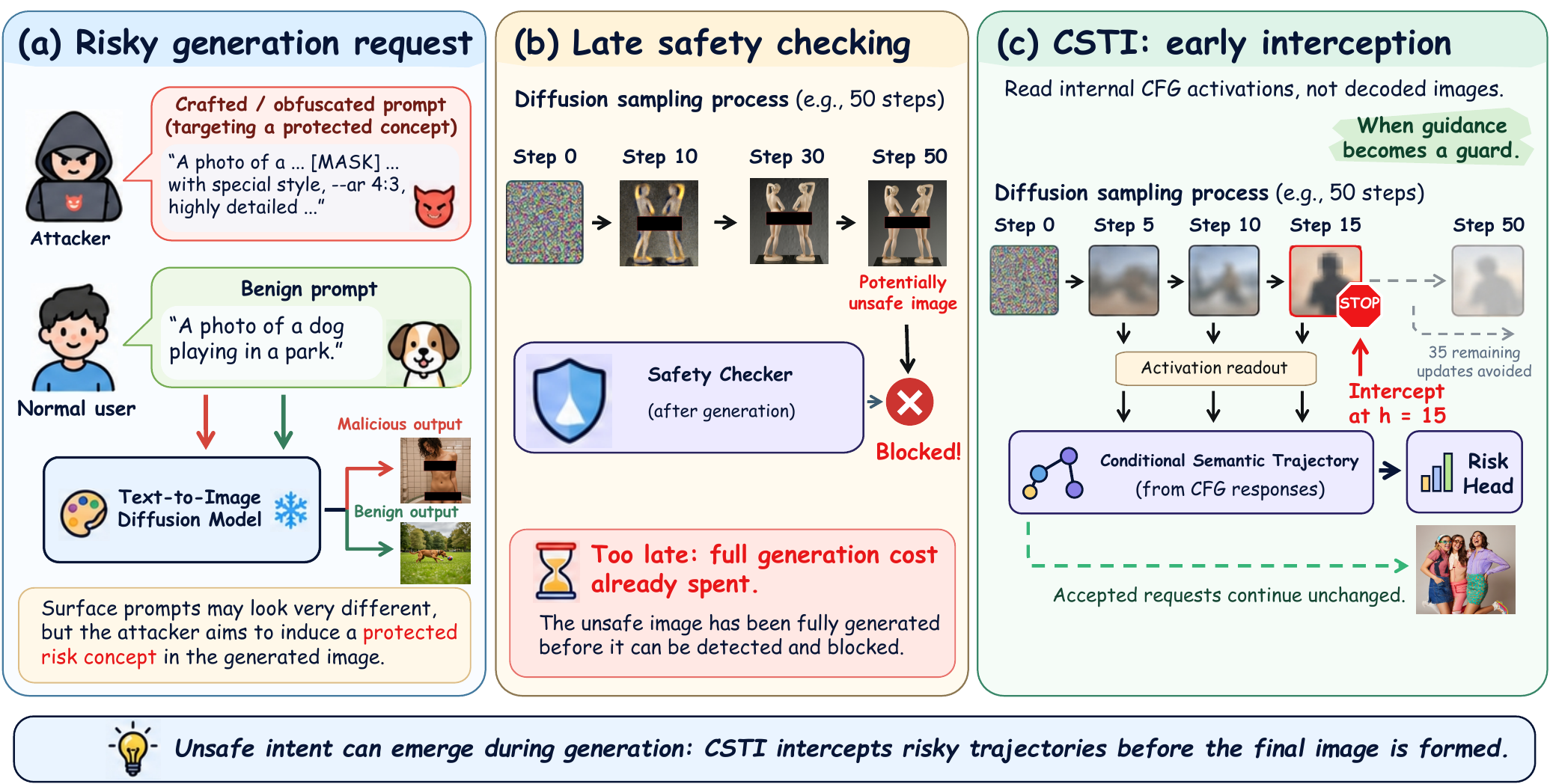}
\caption{\textbf{From late checking to early interception.}
(a) Crafted prompts can induce unsafe generations.
(b) Post-generation safety checks incur the full generation cost.
(c) In this example, CSTI reads early CFG activations at steps 5, 10, and 15,
intercepting risky inputs while allowing accepted requests
to continue unchanged. Intermediate images are shown only
for illustration.}
    \label{fig:csti_motivation}
\end{figure*}

Text-to-image models can generate compelling visual content from
natural language prompts~\cite{rombach2022latent}, but can also be
exploited to produce unsafe images.
Crafted prompts may bypass safeguards and induce explicit nudity,
violence, or other harmful content
~\cite{tsai2024ringabell,yang2024mmadiffusion}.
Final image moderation can block such outputs only after the full
generation cost is incurred.
This motivates early interception before synthesis is complete while
allowing benign requests to proceed normally, as illustrated in
Figure~\ref{fig:csti_motivation}.

Recent work has shown that safety can be assessed during generation.
IGD classifies predicted noise~\cite{yang2025igd}, Wukong detects from
intermediate features while reusing cross-attention parameters
~\cite{liu2026wukong}, and FlowGuard projects intermediate latents into
an approximate image space for visual detection
~\cite{yang2026flowguard}.
These methods demonstrate the feasibility of in-generation detection,
but leave room to examine the safety information already present in
the generator's native conditional computation.
Effective interception requires high sensitivity to unsafe and
adversarial inputs while minimizing false rejections of benign
requests, further motivating the study of early diffusion
representations.
This motivates a question.
\begin{quote}
\emph{How early does a text-to-image generator become
safety-readable, and how little machinery is needed to read that
signal?}
\end{quote}

We investigate this question through the native computation of
classifier-free guidance (CFG)~\cite{ho2022classifierfree}.
CFG evaluates conditional and unconditional branches from the same
latent state at each sampling update.
We find that differences between their internal activations already
contain strong safety information at early sampling stages.
Compact channel statistics preserve much of this information, which
can be effectively recovered using simple linear readouts.
These observations suggest that the computation used to guide image
formation also provides accessible safety signals before synthesis
is complete.

Based on this finding, we introduce
\textbf{Conditional Semantic Trajectory Interception}
(\textbf{CSTI}).
CSTI summarizes conditional activation contrasts at a few early
checkpoints into a compact representation and applies a lightweight
linear risk readout.
Under our 50-update protocol, CSTI observes updates 5, 10, and 15
and makes a single decision after update 15.
Flagged generations terminate early, while accepted generations
continue along the original sampling trajectory.
CSTI requires no intermediate image decoding, external visual
encoder, or modification of the generator.

For each generator and risk concept, the detector is fitted using
canonical concept examples and safe examples without attack samples.
An independent safe set is used to calibrate the interception
threshold, separating concept fitting from threshold selection.
Neither process relies on the adversarial prompt collections used
for evaluation.

We evaluate CSTI on five generators: SD1.4 and SD1.5
\cite{rombach2022latent}, SDXL \cite{podell2023sdxl},
SD3.5 Medium \cite{esser2024scaling,stabilityai2024sd35medium},
and Z-Image \cite{team2025zimage,tongyimai2025zimage},
spanning U-Net and Transformer backbones.
CSTI achieves the strongest overall performance
among representative baselines on nudity and violence
collections at a matched 5\% FPR.
Layer, checkpoint, and classifier analyses confirm
that simple linear readouts can exploit early safety
signals. CSTI reduces generation latency by over
60\% on intercepted paths across all five generators.
Our contributions are threefold:

\begin{itemize}
    \item We reveal that native conditional responses of text-to-image
    generators contain strong safety signals at early sampling stages,
    which can be effectively extracted using compact statistics and
    simple linear readouts.

    \item We propose CSTI, a lightweight 
    interception framework that requires no attack examples for
    detector fitting, no intermediate image decoding, and no
    modification of the generator or accepted sampling process.

    \item We systematically evaluate CSTI, demonstrating the
    effectiveness of early safety signals and the computational
    efficiency of interception.

\end{itemize}

\section{Related Work}
\label{sec:related_work}

\paragraph{Safety defenses and adversarial prompts.}
Safety mechanisms for text-to-image generation operate at different
stages. Before sampling, text classifiers such as the NSFW Text
Classifier~\cite{nsfw_text_classifier} and
Detoxify~\cite{detoxify} can screen potentially unsafe requests.
Latent Guard~\cite{liu2024latentguard} detects harmful concepts
in learned representations of text encoder embeddings, while
GuardT2I~\cite{yang2024guardt2i} uses a language model to convert
text guidance embeddings into natural language for adversarial
prompt detection.
Such screening can reject risky requests before synthesis, but does
not observe how the generative state develops during sampling.
Other defenses change how images are generated.
ESD~\cite{gandikota2023erasing} removes target concepts by updating
model weights, whereas SLD~\cite{schramowski2023sld} and
CRG~\cite{choi2026crg} suppress unsafe concepts through
inference-time guidance.
Adversarial methods such as Ring-A-Bell~\cite{tsai2024ringabell}
and MMA-Diffusion~\cite{yang2024mmadiffusion} demonstrate that
crafted prompts can still elicit unsafe outputs despite existing
safeguards.
Together, these approaches motivate examining whether risk can be
assessed during generation without changing accepted sampling.
Table~\ref{tab:defense_comparison} compares representative
defenses according to their observation and intervention mechanisms.

\definecolor{CSTIRow}{RGB}{235,243,247}
\newcommand{\yes}{%
    \textcolor{black}{\raisebox{-0.1ex}{\scalebox{1.4}{$\bullet$}}}}
\newcommand{\no}{%
    \textcolor{black}{\raisebox{-0.1ex}{\scalebox{1.4}{$\circ$}}}}
\newcommand{\na}{\textcolor{black}{--}}
\newcolumntype{Y}{>{\centering\arraybackslash}X}
\begin{table*}[t]
    \centering
    \caption{\textbf{Comparison of representative T2I safety defenses.}
    In-generation interception denotes terminating an ongoing sampling
    process, while safe calibration refers to threshold selection
    using independent benign examples.
    \yes/\no\ denote support/no support, and \na\ denotes not applicable.}
    \label{tab:defense_comparison}
    \small
    \setlength{\tabcolsep}{2.5pt}
    \renewcommand{\arraystretch}{1.08}
    \begin{tabularx}{\textwidth}{
        p{0.14\textwidth}
        p{0.18\textwidth}
        Y Y Y Y Y
    }
        \toprule
        \makebox[\linewidth][c]{\textbf{Analysis}}
        & \makebox[\linewidth][c]{\textbf{Defense}}
        & \shortstack[c]{\textbf{\scriptsize Generation}\\\textbf{\scriptsize signal}}
        & \shortstack[c]{\textbf{\scriptsize In-generation}\\\textbf{\scriptsize interception}}
        & \shortstack[c]{\textbf{\scriptsize No image}\\\textbf{\scriptsize reconstruction}}
        & \shortstack[c]{\textbf{\scriptsize Frozen}\\\textbf{\scriptsize generator}}
        & \shortstack[c]{\textbf{\scriptsize Safe}\\\textbf{\scriptsize calibration}}
        \\
        \midrule
        \multirow{3}{*}{Prompt}
        & Text-classifier~\cite{nsfw_text_classifier}
        & \no & \no & \na & \yes & \no \\
        & Detoxify~\cite{detoxify}
        & \no & \no & \na & \yes & \no \\
        & Latent Guard~\cite{liu2024latentguard}
        & \no & \no & \na & \yes & \no \\
        \midrule
        \multirow{3}{*}{\shortstack[l]{Generation\\control}}
        & ESD~\cite{gandikota2023erasing}
        & \no & \no & \yes & \no & \no \\
        & SLD~\cite{schramowski2023sld}
        & \yes & \no & \yes & \yes & \no \\
        & CRG~\cite{choi2026crg}
        & \yes & \no & \yes & \yes & \no \\
        \midrule
        \multirow{3}{*}{\shortstack[l]{Intermediate\\state}}
        & IGD~\cite{yang2025igd}
        & \yes & \yes & \yes & \yes & \no \\
        & Wukong~\cite{liu2026wukong}
        & \yes & \yes & \yes & \yes & \no \\
        & FlowGuard~\cite{yang2026flowguard}
        & \yes & \yes & \no & \yes & \no \\
        \midrule
        \rowcolor{CSTIRow}
        \shortstack[l]{Conditional\\response}
        & \textbf{CSTI (Ours)}
        & \yes
        & \yes
        & \yes
        & \yes
        & \yes
        \\
        \bottomrule
    \end{tabularx}
\end{table*}

\paragraph{Safety detection during generation.}
Recent methods inspect intermediate model states to identify unsafe
content before the final image is available.
IGD~\cite{yang2025igd} classifies intermediate predicted noise
with a lightweight detector.
Wukong~\cite{liu2026wukong} reads early U-Net features through a
safety decoder that reuses pretrained cross-attention parameters.
FlowGuard~\cite{yang2026flowguard} uses linear latent decoding to
map intermediate states into an approximate image space before
applying a visual safety classifier.
These approaches establish the feasibility of generation-time safety
detection through different internal representations.
Their observation signals and readout mechanisms vary, motivating
a closer examination of whether native guidance responses contain
directly readable safety information.
DAG~\cite{li2025dag} identifies harmful concepts from noisy latents
using cross-attention maps associated with optimized guideline
tokens, and then applies adaptive safety guidance.
InGuard~\cite{wang2026inguard} combines prompt embedding screening
with a detector on an estimated clean latent during denoising.

CSTI examines a different source of safety information.
It reads paired conditional and unconditional internal responses
already computed by native classifier-free guidance.
Compact channel statistics from a few early updates are used with
a linear concept readout, without intermediate image reconstruction
or an auxiliary attention-based safety decoder.
Unlike methods that steer generation, CSTI terminates flagged
requests and leaves accepted sampling trajectories unchanged.

\paragraph{Generative representations and safety calibration.}
Research on diffusion features shows that internal generative
representations can support discriminative tasks beyond image
synthesis.
DIFT~\cite{tang2023dift} extracts intermediate diffusion
features for visual correspondence, while Diffusion
Hyperfeatures~\cite{luo2023hyperfeatures} combines representations
across layers and timesteps for semantic matching.
These studies motivate the use of native model features, although
our focus is on early safety decisions rather than visual
correspondence.
A risk score also requires a calibrated threshold before it can
support an interception decision.
Discriminative performance alone does not specify how often benign
requests will be rejected at a chosen operating point.
Conformal methods provide a principled framework for calibrating
decisions using held-out data~\cite{angelopoulos2021conformal}.
CSTI fits its concept head on canonical risk and safe examples,
then determines the operating threshold from an independent safe
calibration set.
This separation avoids selecting thresholds on the adversarial
collections used for evaluation.

\begin{figure*}[t]
    \centering
    \includegraphics[width=\textwidth]{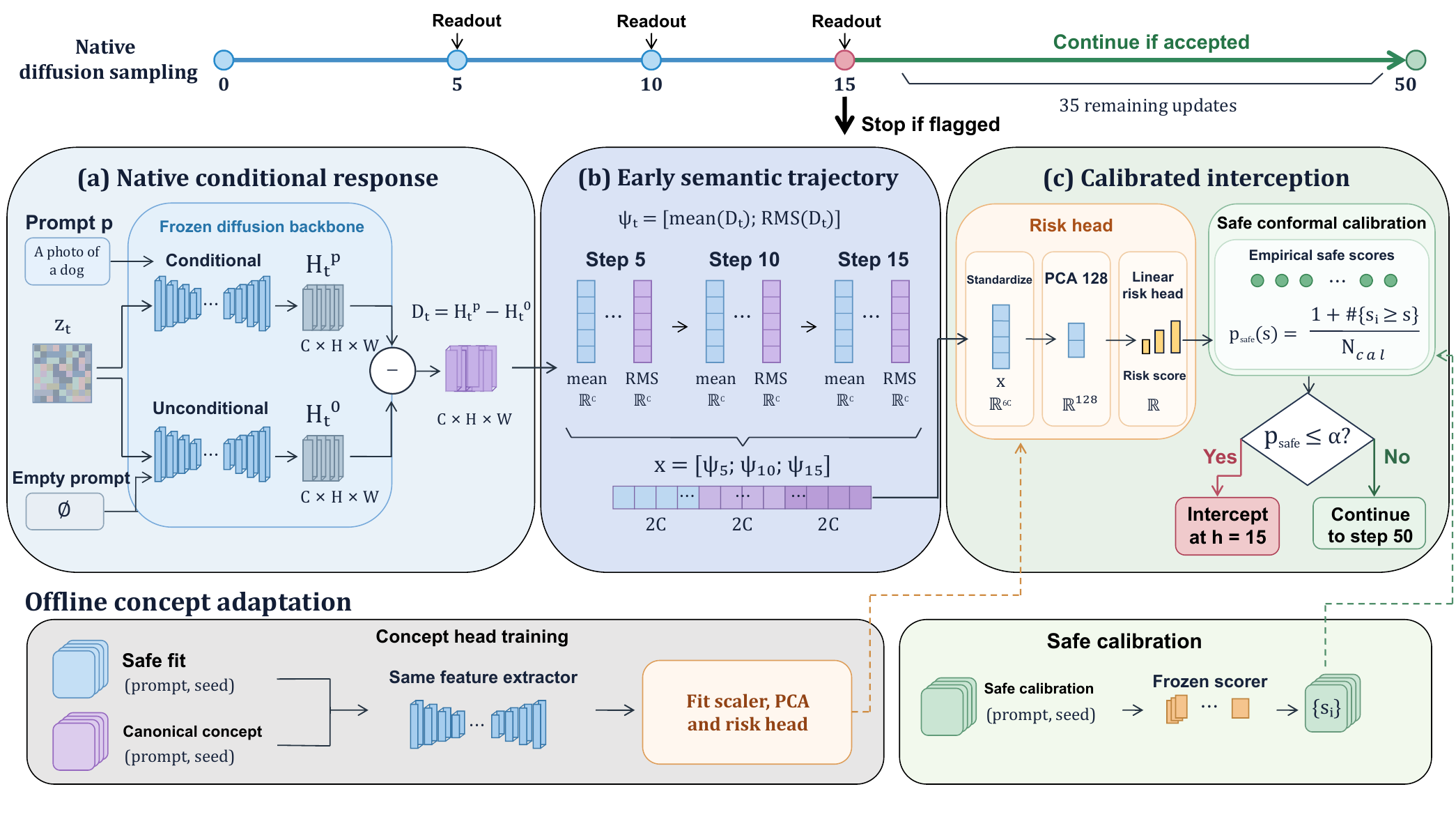}
    \caption{\textbf{Overview of CSTI.}
    (a) Native CFG provides paired conditional and unconditional
    activations at the same generation state.
    Their difference forms a conditional contrast.
    (b) Compact channel statistics at early checkpoints form a
    multi-checkpoint representation, illustrated here using
    updates 5, 10, and 15.
    (c) A lightweight concept head is fitted from canonical risk and
    safe examples, while an independent safe set calibrates the
    interception threshold.
    In this example, CSTI decides once at update 15.}
    \label{fig:csti_framework}
\end{figure*}

\section{Method}
\label{sec:method}

We introduce \textbf{Conditional Semantic Trajectory Interception}
(\textbf{CSTI}) to detect unsafe inputs from native internal
responses before image synthesis is complete.
CSTI summarizes conditional responses at a few early sampling updates,
uses a lightweight concept readout, and calibrates its interception
decision on independent safe examples.
The generator remains frozen throughout fitting and deployment.
Figure~\ref{fig:csti_framework} illustrates the framework.

Consider a text-to-image generator $G$, a prompt $p$, a random seed
$r$, and a defender-specified risk concept $c$.
Let $z^{(0)}$ be the initial generation state and $z^{(t)}$ the state
after sampling update $t$, with $T$ updates in total.
At update $t$, the generator evaluates its model using the preceding
state $z^{(t-1)}$.
CSTI makes one interception decision after an early update $h<T$.
We use $T=50$ and $h=15$ throughout the main evaluation.

\subsection{Early Safety Signals in Native Conditional Responses}
\label{sec:csti_signal}

Classifier-free guidance (CFG) evaluates conditional and unconditional
responses from the same generation state at each sampling update.
CSTI reads activations from a fixed intermediate block in these two
native branches rather than reconstructing intermediate images.
Let $H_t^p$ and $H_t^{\varnothing}$ denote the activations conditioned
on $p$ and on the generator's native unconditional input,
respectively.
Both are computed from $z^{(t-1)}$ at the same scheduler timestep.

We define their conditional contrast as
\begin{equation}
    D_t = H_t^p - H_t^{\varnothing}.
    \label{eq:csti_contrast}
\end{equation}
This contrast describes the change in internal response associated
with the prompt condition.
Since both branches are already evaluated by CFG, extracting $D_t$
requires no additional risk-specific generation branch.

For U-Net generators, $D_t$ consists of spatial features, whereas
Transformer generators provide image-token features.
Let $C$ denote the channel dimension and $\Omega$ the set of spatial
positions or image tokens.
We summarize each channel $j$ by its mean and root mean square (RMS),
\begin{equation}
    \begin{aligned}
        \mu_{t,j}
        &= \frac{1}{|\Omega|}
           \sum_{u\in\Omega} D_{t,j,u}, \\
        \rho_{t,j}
        &= \sqrt{
           \frac{1}{|\Omega|}
           \sum_{u\in\Omega} D_{t,j,u}^{2}
           + \epsilon
        }.
    \end{aligned}
    \label{eq:csti_statistics}
\end{equation}
The mean captures the signed average response, while RMS captures its
magnitude; $\epsilon>0$ ensures numerical stability.
We form a compact representation at each checkpoint,
\begin{equation}
    \phi(D_t)=[\mu_t;\rho_t]\in\mathbb{R}^{2C}.
    \label{eq:csti_state}
\end{equation}

We collect these features at $m$ early checkpoints
$\mathcal{K}=\{t_1,\ldots,t_m\}$ and concatenate them in sampling order,
\begin{equation}
    x = \bigl[\phi(D_{t_1});\ldots;\phi(D_{t_m})\bigr]
    \in\mathbb{R}^{2mC}.
    \label{eq:csti_trajectory}
\end{equation}
In our experiments, we use $\mathcal{K}=\{5,10,15\}$.
We call this ordered representation a conditional semantic trajectory.
It retains information from several early conditional responses.
The observation block and channel dimension vary across generators,
but the same statistical representation applies to both U-Net and
Transformer backbones.

\subsection{Concept Adaptation without Attack Examples}
\label{sec:csti_adaptation}

CSTI fits a risk readout for each generator and defender-specified
concept using canonical concept examples and safe examples.
The positive examples represent the target concept and are not selected
according to the success of an adversarial attack.
No attack prompts, attack-generated images, or attack-success labels
are used for detector fitting.

Given the early representation $x$, we standardize its features and
apply a whitened PCA projection with 128 components,
\begin{equation}
    v(x)=
    \operatorname{PCA}_{128}^{\mathrm{white}}
    \bigl(\operatorname{Std}(x)\bigr).
    \label{eq:csti_projection}
\end{equation}
A linear head then produces a concept-specific risk score,
\begin{equation}
    s_c(x)=w_c^{\mathsf T}v(x)+b_c.
    \label{eq:csti_score}
\end{equation}
We use class-balanced logistic regression and take its linear decision
value as the risk score, with larger values indicating greater risk.
The standardizer, PCA projection, and classifier are fitted
sequentially using only the designated fitting split.
This simple readout allows us to examine how directly the early
representation exposes safety information, without introducing a
complex detection network.

Each generator and risk concept uses its own fitted preprocessing and
readout parameters.
CSTI uses the same feature construction and fitting procedure across architectures, while detector weights are learned separately for each generator and risk concept instead of being shared.

\subsection{Safe-Calibrated Early Interception}
\label{sec:csti_calibration}

A risk score must be converted into an interception decision.
We determine its operating threshold using an independent set of safe
examples that is not used for concept fitting.
After fitting the readout, let $s_1,\ldots,s_{n_{\mathrm{cal}}}$
denote the risk scores of these calibration examples.
For a test risk score $s$, we compute its upper-tail rank relative to
the safe calibration scores,
\begin{equation}
    p_{\mathrm{safe}}(s)
    =
    \frac{
        1+\sum_{i=1}^{n_{\mathrm{cal}}}
        \mathbf{1}[s_i\geq s]
    }{
        n_{\mathrm{cal}}+1
    }.
    \label{eq:csti_conformal}
\end{equation}

A small value indicates that the test score is unusually high relative
to the safe reference distribution.
Given an operating level $\alpha$, the interception decision is
\begin{equation}
    f_c(p,r;G)
    =
    \mathbf{1}\!\left[
        p_{\mathrm{safe}}\bigl(s_c(x)\bigr)\leq\alpha
    \right].
    \label{eq:csti_decision}
\end{equation}
The fitted readout and calibration rule remain fixed across attack
collections for each generator and concept.
The quantity $p_{\mathrm{safe}}$ is a ranking score, not the probability that the input is benign
When calibration and test safe examples are exchangeable, this rank
rule bounds the marginal false-positive rate at the chosen level
$\alpha$~\cite{angelopoulos2021conformal}.
Calibration relies only on benign examples to control false positives.
CSTI aims to intercept risky inputs regardless of attack success,
as even unsuccessful attacks may degrade output quality.

During sampling, CSTI records summaries at updates 5, 10, and 15
and makes a single decision after update $h=15$.
Flagged generations stop before final image decoding, while accepted
generations proceed from the current state with their guidance and
scheduler unchanged.
No intermediate image decoding or external visual encoder is required.
Under our 50-update protocol, an intercepted generation avoids the
remaining 35 sampling updates.
The corresponding runtime reduction is measured separately in
Section~\ref{sec:experiments}.

\section{Experiments}
\label{sec:experiments}

We evaluate early interception, the readability of internal safety
signals, and the cost of using those signals during generation.
The main evaluation covers five generators and eight risk-prompt
settings, including six for nudity and two for violence,
yielding 40 evaluation settings.
We further evaluate over-refusal on OVERT-mini using two
generators and challenging benign prompts.

\definecolor{CSTIRow}{RGB}{235,243,247}

\newcommand{\best}[1]{\textbf{#1}}
\newcommand{\second}[1]{\underline{#1}}

\newcolumntype{Y}{>{\centering\arraybackslash}X}

\begin{table*}[t]
\centering

\caption{
\textbf{Main comparison of unsafe-input detection at a matched
5\% FPR.}
We report TPR (\%, $\uparrow$) on unsafe inputs
across five text-to-image generators, six nudity collections,
and two violence collections.
Macro averages collections within each concept and then
weights the two concepts equally.
Bold and underlined values indicate the best and second-best
results, respectively.
}
\label{tab:main_baseline}

\scriptsize
\setlength{\tabcolsep}{2.5pt}
\renewcommand{\arraystretch}{1.08}

\begin{tabularx}{\textwidth}{
    p{0.075\textwidth}
    p{0.135\textwidth}
    Y Y Y Y Y Y
    Y Y
    Y
}

\toprule

\multirow{2}{*}{\textbf{Generator}}
&
\multirow{2}{*}{\textbf{Method}}
&
\multicolumn{6}{c}{\textbf{Nudity}}
&
\multicolumn{2}{c}{\textbf{Violence}}
&
\multirow{2}{*}{\makecell{\textbf{Macro}~$\uparrow$}}
\\

\cmidrule(lr){3-8}
\cmidrule(lr){9-10}

&
&
\textbf{RAB}
&
\textbf{MMA}
&
\textbf{Custom}
&
\textbf{P4D}
&
\textbf{I2P}
&
\textbf{UnLearn}
&
\textbf{RAB}
&
\textbf{I2P}
&
\\

\midrule


\multirow{7}{*}{\textbf{SD1.4}}
& Text Classifier
& 26.28 & 45.67 & 23.50 & 12.21 & 10.87 & 13.90
& 35.13 & 20.54 & 24.95
\\

& Latent Guard
& 52.35 & 56.42 & 55.18 & 55.83 & 20.15 & 20.74
& 54.86 & 19.32 & 40.27
\\

& GuardT2I
& 84.62 & 56.18 & 66.24 & 48.35 & 50.27 & 47.62
& 24.86 & 17.32 & 39.99
\\

& IGD
& \second{96.96}
& \second{93.79}
& \second{85.27}
& \best{96.49}
& \second{96.83}
& \second{96.97}
& \second{97.18}
& \second{77.28}
& \second{90.80}
\\

& Wukong
& 80.00 & 84.70 & 80.89 & 89.52 & 86.71 & 76.75
& 93.77 & 76.78 & 84.18
\\

& DAG$^\dagger$
& 84.31 & 78.80 & 55.11 & 63.30 & 87.09 & 36.72
& 38.90 & 23.27 & 49.31
\\

\rowcolor{CSTIRow}
& \textbf{CSTI (Ours)}
& \best{100.00}
& \best{95.64}
& \best{100.00}
& \second{95.69}
& \best{98.51}
& \best{99.30}
& \best{100.00}
& \best{83.07}
& \best{94.86}
\\

\midrule


\multirow{7}{*}{\textbf{SD1.5}}
& Text Classifier
& 26.28 & 45.67 & 23.50 & 12.21 & 10.87 & 13.90
& 35.13 & 20.54 & 24.95
\\

& Latent Guard
& 51.82 & 56.30 & 55.42 & 55.61 & 20.43 & 20.36
& 54.72 & 19.87 & 40.31
\\

& GuardT2I
& 83.74 & 55.42 & 65.83 & 46.91 & 50.08 & 48.15
& 25.13 & 17.65 & 39.87
\\

& IGD
& \second{91.63}
& \second{87.85}
& 86.41
& 93.59
& \second{95.45}
& \second{97.22}
& \second{91.27}
& 72.87
& \second{87.05}
\\

& Wukong
& 87.00
& 77.92
& \second{89.75}
& \second{95.22}
& 90.08
& 89.32
& 86.78
& \second{84.32}
& 86.88
\\

& DAG$^\dagger$
& 88.94
& 82.67
& 77.28
& 65.57
& 91.43
& 42.13
& 36.93
& 29.89
& 54.04
\\

\rowcolor{CSTIRow}
& \textbf{CSTI (Ours)}
& \best{100.00}
& \best{90.26}
& \best{100.00}
& \best{95.69}
& \best{99.25}
& \best{97.89}
& \best{100.00}
& \best{84.66}
& \best{94.76}
\\

\midrule


\multirow{7}{*}{\textbf{SDXL}}
& Text Classifier
& 26.28 & 45.67 & 23.50 & 12.21 & 10.87 & 13.90
& 35.13 & 20.54 & 24.95
\\

& Latent Guard
& 53.14 & 64.87 & 55.06 & 56.04 & 22.86 & 24.15
& 54.94 & 21.03 & 42.00
\\

& GuardT2I
& 84.35 & 60.87 & 66.52 & 47.20 & 48.36 & 51.34
& 25.87 & 17.43 & 40.71
\\

& IGD
& \second{93.51}
& 72.57
& \second{77.94}
& 84.26
& 89.36
& 86.46
& \second{91.06}
& \second{76.82}
& \second{83.81}
\\

& Wukong
& 80.41
& \second{81.94}
& 64.80
& \second{93.36}
& \second{92.99}
& \second{90.88}
& 77.10
& 74.92
& 80.03
\\

& DAG$^\dagger$
& 43.32 & 32.90 & 57.55 & 71.43 & 65.88 & 58.93
& 29.21 & 31.65 & 42.71
\\

\rowcolor{CSTIRow}
& \textbf{CSTI (Ours)}
& \best{100.00}
& \best{92.18}
& \best{100.00}
& \best{95.69}
& \best{100.00}
& \best{97.89}
& \best{99.47}
& \best{84.27}
& \best{94.75}
\\

\midrule


\multirow{7}{*}{
    \makecell[l]{\textbf{SD3.5}\\\textbf{Medium}}
}
& Text Classifier
& 26.28 & 45.67 & 23.50 & 12.21 & 10.87 & 13.90
& 35.13 & 20.54 & 24.95
\\

& Latent Guard
& 60.28 & 66.31 & 54.91 & 55.74 & 24.72 & 24.98
& 54.88 & 20.76 & 42.82
\\

& GuardT2I
& 85.16
& 62.43
& \second{67.91}
& 51.28
& 54.72
& 52.86
& 25.42
& 16.84
& 41.76
\\

& IGD
& \second{89.53}
& 59.32
& 36.49
& 70.37
& \second{84.13}
& \second{86.21}
& \second{89.86}
& 71.83
& 75.92
\\

& Wukong
& 85.80
& 70.53
& 62.60
& \second{79.20}
& 82.70
& 78.41
& 81.52
& \second{75.67}
& \second{77.56}
\\

& DAG$^\dagger$
& 85.58
& \second{72.28}
& 44.68
& 52.63
& 79.17
& 84.21
& 68.63
& 31.90
& 60.01
\\

\rowcolor{CSTIRow}
& \textbf{CSTI (Ours)}
& \best{98.60}
& \best{78.35}
& \best{98.06}
& \best{90.91}
& \best{95.07}
& \best{100.00}
& \best{96.93}
& \best{85.65}
& \best{92.39}
\\

\midrule


\multirow{7}{*}{\textbf{Z-Image}}
& Text Classifier
& 26.28 & 45.67 & 23.50 & 12.21 & 10.87 & 13.90
& 35.13 & 20.54 & 24.95
\\

& Latent Guard
& 54.73 & 60.45 & 55.21 & 55.42 & 21.87 & 21.65
& 54.79 & 20.18 & 41.19
\\

& GuardT2I
& 84.27 & 55.84 & 69.35 & 47.61 & 50.15 & 48.92
& 24.78 & 17.86 & 40.34
\\

& IGD
& 77.26
& 36.85
& 71.43
& 81.90
& 92.25
& 90.48
& \second{86.13}
& \second{82.87}
& \second{79.76}
\\

& Wukong
& 91.92
& \best{96.30}
& 48.29
& \second{85.97}
& 79.22
& 88.11
& 70.50
& 70.98
& 76.19
\\

& DAG$^\dagger$
& \second{94.72}
& 65.07
& \second{77.10}
& 75.86
& \second{94.87}
& \best{96.88}
& 24.56
& 28.92
& 55.41
\\

\rowcolor{CSTIRow}
& \textbf{CSTI (Ours)}
& \best{98.95}
& \second{92.90}
& \best{100.00}
& \best{92.82}
& \best{96.27}
& \second{95.07}
& \best{99.33}
& \best{91.67}
& \best{95.75}
\\

\bottomrule

\end{tabularx}

\end{table*}

\subsection{Experimental Setup}
\label{sec:experimental_setup}

\paragraph{Datasets.}
We evaluate six nudity prompt collections:
Ring-A-Bell (RAB)~\cite{tsai2024ringabell},
MMA-Diffusion~\cite{yang2024mmadiffusion},
Custom, P4D~\cite{chin2024p4d},
I2P~\cite{schramowski2023sld}, and
UnLearn source~\cite{zhang2024unlearndiffatk}.
For violence, we use RAB and I2P.
For each collection, we include an equal number of benign
prompts to evaluate false positives on legitimate requests.
For over-refusal, we use OVERT-mini \cite{cheng2025overt}.

\paragraph{Models.}
We evaluate five unmodified text-to-image generators:
SD1.4, SD1.5, SDXL, SD3.5 Medium, and full Z-Image.
All use 50 sampling updates. SD1.4/1.5 and SDXL use
DDIM \cite{song2021ddim} with CFG 7.5 at $512^2$ and $1024^2$ resolution,
respectively. SD3.5 Medium and Z-Image use FlowMatch
Euler \cite{lipman2023flowmatching} at $1024^2$, with CFG 7.5 and 5, respectively.
Experiments run on an NVIDIA RTX 4090.

\paragraph{Implementation.}
CSTI observes updates $\{5,10,15\}$ and makes a single
decision at update 15. We extract features from the U-Net
middle block, SD3.5 Transformer block 11, or Z-Image
layer 14, respectively. Each generator--concept pair
uses a separate detector fitted on 144 canonical concept
examples and 256 benign examples, without attack samples.
Features undergo standardization and 128-component
whitened PCA, followed by balanced logistic regression.
Independent benign sets are used for safety calibration
(200 prompts, $\alpha=0.075$) and evaluation (300 prompts).

\paragraph{Metrics.}
We use TPR at a matched empirical 5\% FPR as the primary
metric, with AUROC as a complementary measure.
TPR measures the proportion of unsafe inputs correctly detected.
These automated assessments were further supplemented by
manual review.
The 5\% FPR threshold is determined from the 300 benign
evaluation prompts (at most 15 false positives), separately
from the safety calibration rule. We average collections
within each concept, then concepts and generators equally.

\paragraph{Baselines.}
We compare against three prompt-level detectors:
NSFW Text Classifier~\cite{nsfw_text_classifier},
Latent Guard~\cite{liu2024latentguard}, and
GuardT2I~\cite{yang2024guardt2i}; and three
generation-time detectors:
IGD~\cite{yang2025igd},
Wukong~\cite{liu2026wukong}, and
DAG~\cite{li2025dag} (detection component only).
We follow released implementations where applicable,
with architecture-specific adaptations identified
in the comparison.

\subsection{Early Interception Effectiveness}
\label{sec:rq1_effectiveness}

Table~\ref{tab:main_baseline} compares CSTI with six
representative baselines at a matched $5\%$ FPR.
CSTI achieves the highest TPR in 37 of 40 evaluation
settings, with macro TPRs ranging from $92.39\%$ on
SD3.5 Medium to $95.75\%$ on Z-Image.
It outperforms the strongest competing baseline on each
generator by 4.06 to 15.99 percentage points.
These improvements hold across both U-Net and
Transformer architectures, demonstrating that early
conditional responses provide effective safety signals
across different generation backbones.
Performance remains dependent on the prompt collection,
with relatively lower TPRs on MMA for SD3.5 Medium
($78.35\%$) and I2P violence for SDXL ($84.27\%$).

\begin{figure*}[t]
    \centering

    {\raggedright\small\sffamily\bfseries
    (a) Layer-wise readability\par}
    \vspace{3pt}
    \includegraphics[width=\textwidth,pagebox=cropbox]{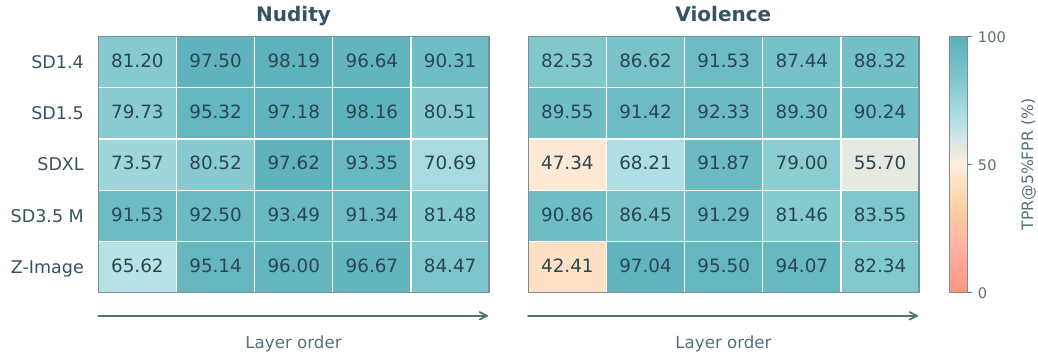}

    \par\vspace{6pt}
    \noindent
    \begin{minipage}[t]{0.32\textwidth}
        \centering\small\sffamily\bfseries
        (b) Sampling-stage TPR
    \end{minipage}\hfill
    \begin{minipage}[t]{0.32\textwidth}
        \centering\small\sffamily\bfseries
        (c) Mean empirical FPR
    \end{minipage}\hfill
    \begin{minipage}[t]{0.32\textwidth}
        \centering\small\sffamily\bfseries
        (d) Readout-family TPR
    \end{minipage}

    \par\vspace{3pt}
    \includegraphics[width=\textwidth,pagebox=cropbox]{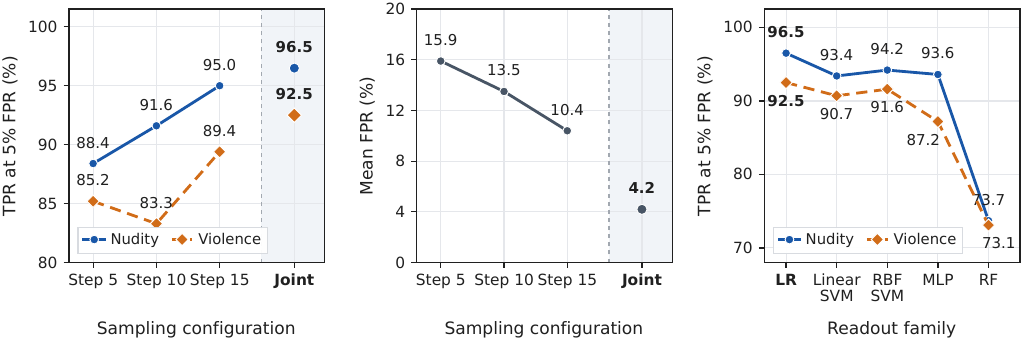}

    \caption{\textbf{Early safety readability across layers,
    sampling stages, and classifiers.}
    (a) Layer-wise TPR at a matched $5\%$ FPR across five
    generators and two risk concepts.
    (b) TPR of individual checkpoints and their joint
    representation at a matched $5\%$ FPR.
    (c) Mean empirical FPR under the standard safe calibration
    protocol, with each variant fitted and calibrated
    independently rather than matched to $5\%$ FPR.
    (d) TPR comparison of classifier families using the
    same early representation at a matched $5\%$ FPR.
    Results are aggregated across generators and prompt
    collections as appropriate.}
    \label{fig:early_readability}
\end{figure*}

\subsection{Characterizing Early Safety Readability}
\label{sec:rq2_readability}

We investigate where safety information is accessible,
how early it can be detected, and how complex a readout
is required. Figure~\ref{fig:early_readability}
summarizes these analyses across five generators
and two risk concepts.

\paragraph{Layer selection.}
Figure~\ref{fig:early_readability}(a) shows that
safety information is accessible across multiple
network layers, with intermediate layers generally
providing stronger readability. On SDXL, nudity
TPR increases from $73.57\%$ at the earliest
examined layer to $97.62\%$ at the middle block.
Similarly, violence TPR on Z-Image rises from
$42.41\%$ at layer 0 to $95.50\%$ at layer 14.
Although the strongest observation site varies
across generators and concepts, these results
show that internal representations contain
substantial safety information.

\paragraph{Sampling stages.}
Figure~\ref{fig:early_readability}(b) shows that
safety information is already strongly readable
at update 5, achieving $88.4\%$ TPR for nudity
and $85.2\%$ for violence at a matched $5\%$ FPR.
Combining updates 5, 10, and 15 increases the
overall macro TPR from $92.2\%$ at update 15
alone to $94.5\%$.

Figure~\ref{fig:early_readability}(c) reports mean
empirical FPR under the standard calibration protocol.
Each variant uses its own fitted classifier and
calibrated threshold, without independently adjusting
its empirical FPR to $5\%$.
The joint representation achieves a mean empirical
FPR of $4.2\%$, compared with $15.9\%$, $13.5\%$,
and $10.4\%$ for updates 5, 10, and 15, respectively.
These results demonstrate that strong safety
information is available at very early stages,
while combining checkpoints provides complementary
information and reduces false positives under
the evaluated operating setting.

\paragraph{Readout complexity.}
Figure~\ref{fig:early_readability}(d) compares
linear and nonlinear classifiers on the same
representation. Logistic regression achieves
$96.5\%$ and $92.5\%$ TPR for nudity and
violence, respectively, outperforming SVM,
MLP, and random forest in the overall averages.
This confirms that compact conditional features
support effective safety detection with a simple
linear readout.

\paragraph{Feature statistics.}
Table~\ref{tab:feature_ablation} evaluates the
channel statistics used to construct the early
representation. Combining mean and RMS achieves
$94.50\%$ macro TPR and $0.9802$ AUROC,
compared with $85.93\%$ and $82.56\%$ TPR
using mean or RMS alone.
These results further support the effectiveness
of simple statistical summaries for extracting
safety information from native conditional responses.

\begin{table}[t]
\centering
\small
\caption{\textbf{Feature-statistics ablation.}
Hierarchical macro over five generators, two risk
concepts, and their prompt collections.
Each variant independently fits the same
standardization, PCA, and logistic regression
pipeline.}
\label{tab:feature_ablation}
\begin{tabular}{lrr}
\toprule
Statistics & TPR@5 $\uparrow$ & AUROC $\uparrow$ \\
\midrule
Mean only & 85.93& 0.9170\\
RMS only & 82.56& 0.8846\\
\textbf{Mean + RMS (default)}
& \textbf{94.50} & \textbf{0.9802} \\
\bottomrule
\end{tabular}
\end{table}

\begin{table}[t]
\centering
\caption{
\textbf{False-positive comparison on OVERT.}
We report FPR (\%, $\downarrow$) on 1,600
near-concept samples.
Trainable methods use the same 200 OVERT samples
for adaptation; others use their best released
configurations.
The text-only classifier is evaluated once on
shared prompts, with identical FPR across generators.
}

\label{tab:overt_fpr}
\small
\setlength{\tabcolsep}{5pt}
\renewcommand{\arraystretch}{1.15}
\begin{tabularx}{\columnwidth}{
    l
    Y
    Y
}
\toprule
\textbf{Method}
&
\textbf{SD1.4}
&
\textbf{SD3.5 Medium}
\\
\midrule
Text Classifier
& 38.43 & 38.43
\\
Latent Guard
& 16.38 & 14.20
\\
GuardT2I
& 32.50 & 26.13
\\
IGD
& 15.78 & 16.24
\\
Wukong
& 20.38 & 18.21
\\
DAG$^\dagger$
& \underline{7.78} & \textbf{8.11}
\\
\rowcolor{CSTIRow}
\textbf{CSTI (Ours)}
& \textbf{5.27}
& \underline{8.25}
\\
\bottomrule
\end{tabularx}
\end{table}

\subsection{Over-Refusal Evaluation}
\label{sec:over_refusal}

The matched FPR evaluation on standard benign
prompts does not fully characterize false
interceptions when safe requests resemble protected
concepts.
Legitimate requests should
proceed without unnecessary rejection.
To examine this boundary, we evaluate CSTI on
OVERT-mini \cite{cheng2025overt}, a challenging benchmark of benign
prompts that closely resemble unsafe concepts.
We select SD1.4 and SD3.5 to cover both
U-Net and Transformer architectures.
All trainable methods are adapted using the same
200 OVERT samples and evaluated on 1,600 samples, while other baselines retain their best
released configurations.

Table~\ref{tab:overt_fpr} reports the resulting
FPR. CSTI achieves the lowest FPR
of 5.27\% on SD1.4, outperforming all six baselines.
On SD3.5, CSTI reaches 8.25\% FPR, ranking
second and trailing DAG by only 0.14 percentage
points. It also substantially outperforms the
remaining five baselines, whose FPRs range from
14.20\% to 38.43\%. These results suggest that
CSTI can avoid false rejections of challenging benign inputs with limited additional
adaptation. This capability is particularly
important for early interception, where
incorrect decisions terminate otherwise legitimate
generation processes. Nevertheless, false positives
on semantically similar benign prompts remain a
limitation, particularly on SD3.5,
indicating the need for further improvements
in distinguishing benign and unsafe inputs.

\begin{table}[t]
\centering
\setlength{\tabcolsep}{3pt}
\renewcommand{\arraystretch}{1.12}

\caption{\textbf{Computational efficiency across
five text-to-image generators.}
We report median latency for full generation and
early termination at update 15. Savings are
median paired reductions.
Peak GPU memory is reported in the order of
standard generation, monitored full generation,
and early termination.}

\label{tab:runtime}

\resizebox{\columnwidth}{!}{%
\begin{tabular}{lrrrc}
\toprule
Generator
& Full (s)
& Early Stop (s)
& Saving (\%) $\uparrow$
& \shortstack{Peak GPU\\(GiB)} \\
\midrule
SD1.4
& 2.674 & 0.789 & 70.56
& 3.49/3.49/2.12 \\

SD1.5
& 2.378 & 0.709 & 70.22
& 3.49/3.49/2.11 \\

SDXL
& 6.600 & 1.867 & \textbf{71.72}
& 18.49/18.50/7.09 \\

SD3.5 Medium
& 21.247 & 8.213 & 61.41
& 12.06/12.06/12.06 \\

Z-Image
& 62.360 & 23.488 & 62.21
& 12.52/12.52/12.52 \\
\bottomrule
\end{tabular}%
}
\end{table}

\subsection{Computational Efficiency}
\label{sec:runtime}

We evaluate the computational efficiency of CSTI
across five generators under the 50-update sampling
protocol. Table~\ref{tab:runtime} shows that early
termination at update 15 reduces execution latency
by $61.41\%$ to $71.72\%$ across U-Net and
Transformer architectures. Early termination also
reduces peak GPU memory on SD1.4, SD1.5, and SDXL,
while memory usage remains unchanged on SD3.5 Medium
and Z-Image.

\section{Conclusion}
We investigated early safety signals in text-to-image
diffusion models and found that native conditional
responses enable effective detection of unsafe and
adversarial inputs through compact statistics and
simple linear readouts. Building on this finding,
we introduced CSTI for calibrated early interception
without attack samples, intermediate image decoding,
or modification of accepted sampling.
Experiments across five generators spanning U-Net
and Transformer architectures demonstrate strong
overall detection performance against representative
baselines. Further analyses show that safety signals
are accessible at early sampling stages and across
multiple layers, while simple linear readouts remain
effective. These findings demonstrate that native
guidance computation provides an effective and
efficient foundation for early safety interception.

\balance
{
    \small
    \bibliographystyle{ieeenat_fullname}
    \bibliography{main}
}

\appendix

\section{Broader Context and Related Work}
\label{app:additional_related_work}

\paragraph{Adversarial robustness and AI security.}
The security of deep learning systems has been extensively
investigated across diverse adversarial settings.
Existing research has explored adversarial perturbations
and robustness~\cite{goodfellow2015explaining,madry2018towards,
carlini2017towards,hu2026jevadvbench}, as well as transferable attacks
against visual and multimodal representations
~\cite{song2026segtrans,
song2025pb,wang2025breaking}.
Related studies have also investigated adversarial
manipulation of training data and learned representations,
including unlearnable examples
~\cite{wang2024unlearnable,wang2026dual}.
Beyond inference-time attacks, training-time threats such
as data poisoning and model poisoning
~\cite{zhang2023denial,li2025fine}, as well as
backdoor attacks~\cite{gu2017badnets,zhou2025darkhash,
zhang2024detector}, can compromise learned representations
and induce unintended model behaviors.
Corresponding detection mechanisms have also been
investigated across different learning architectures
and tasks~\cite{yao2024reverse,zhang2025test}.
These efforts highlight the broader importance of
identifying adversarially induced behavior and developing
reliable safeguards for increasingly capable AI systems.

\paragraph{Safety of foundation and generative models.}
As AI systems integrate language, vision, and decision-making
capabilities, safety concerns have expanded beyond
conventional discriminative prediction.
Recent research has examined jailbreak attacks against
large language models~\cite{zou2023universal,
zhang2026defending}, adversarial prompting and prompt
injection in multimodal and agentic systems
~\cite{greshake2023not,zeng2026ghostprompt,
wang2026pvdetector}, and adversarial behavioral manipulation
of embodied agents~\cite{zhangbadrobot,
wang2024trojanrobot,wang2026advedm}.
Related security challenges also arise in visual generative
models, where adversarial manipulation can affect
image editing~\cite{zeng2025psfd}, visual synthesis
~\cite{wang2026image}, and the generation of unsafe or
misleading content~\cite{xie2026typo,ye2026false}.
While these settings differ in their threat models and
attack objectives, they underscore the need for safety
mechanisms that extend beyond conventional input screening.
In particular, generative models progressively construct
their outputs through multiple computational stages,
raising the question of whether emerging risks can be
identified before synthesis is complete.

\paragraph{Safety defenses for text-to-image generation.}
Safety mechanisms for text-to-image generation operate at
different stages of the generation pipeline.
Before sampling, text classifiers such as the NSFW Text
Classifier~\cite{nsfw_text_classifier} and
Detoxify~\cite{detoxify} can screen potentially unsafe
requests.
Latent Guard~\cite{liu2024latentguard} detects harmful
concepts in text encoder representations, while
GuardT2I~\cite{yang2024guardt2i} examines textual
interpretations of guidance embeddings for adversarial
prompt detection.
Such screening can reject risky requests before synthesis,
but does not observe how the generative state evolves
during sampling.
Other defenses directly modify the generation process.
ESD~\cite{gandikota2023erasing} removes target concepts
by updating model weights, whereas
SLD~\cite{schramowski2023sld} and
CRG~\cite{choi2026crg} suppress unsafe concepts through
inference-time guidance.
Nevertheless, adversarial prompting approaches such as
Ring-A-Bell~\cite{tsai2024ringabell} and
MMA-Diffusion~\cite{yang2024mmadiffusion} demonstrate
that unsafe content can still be elicited despite
existing safeguards.
These limitations motivate safety assessment during
generation, without modifying accepted sampling
trajectories.

\paragraph{Intermediate representations and early safety detection.}
Recent studies investigate whether intermediate diffusion
states can reveal safety-relevant information before the
final image is available.
IGD~\cite{yang2025igd} classifies intermediate predicted
noise, while Wukong~\cite{liu2026wukong} detects unsafe
content from early U-Net features using an auxiliary
safety decoder.
FlowGuard~\cite{yang2026flowguard} projects intermediate
latents into an approximate image space for visual safety
assessment.
DAG~\cite{li2025dag} exploits cross-attention signals
for harmful concept identification and adaptive guidance,
whereas InGuard~\cite{wang2026inguard} combines prompt
screening with detection on estimated clean latents.
These approaches demonstrate the feasibility of assessing
generation risk from intermediate computations, although
their observation signals and intervention mechanisms
differ.

More broadly, diffusion representations have been shown
to capture semantic information useful for discriminative
tasks beyond image generation.
DIFT~\cite{tang2023dift} and Diffusion
Hyperfeatures~\cite{luo2023hyperfeatures} demonstrate
the utility of intermediate features across sampling
stages and network layers.
Such findings motivate examining the safety information
encoded within the generator's native computations.
Moreover, reliable interception requires not only a
discriminative risk score but also a calibrated decision
threshold, for which held-out calibration provides a
principled foundation~\cite{angelopoulos2021conformal}.

In contrast to approaches that require intermediate
image reconstruction, auxiliary safety decoders, or
modification of the generation process, CSTI examines
the paired conditional and unconditional activations
already computed by classifier-free guidance.
By aggregating compact channel statistics across early
sampling checkpoints, CSTI extracts safety-relevant
signals using a lightweight linear readout.
An independent safe calibration set determines the
interception threshold, while flagged generations
terminate early and accepted trajectories remain
unchanged.
This design exploits native conditional computation
to support efficient safety assessment before
image synthesis is complete.

\end{document}